%% file: main.tex
\documentclass[11pt,letterpaper]{article}
\usepackage[letterpaper]{geometry}
\usepackage{aaai18}
\usepackage{times}

\usepackage{latexsym}
\usepackage{amsmath}
\usepackage{multirow}
\usepackage{url}
\makeatletter
\newcommand{\@BIBLABEL}{\@emptybiblabel}
\newcommand{\@emptybiblabel}[1]{}
\makeatother
\DeclareMathOperator*{\argmax}{arg\,max}

\usepackage{enumitem}
\usepackage{graphicx}
\usepackage{subfigure}
\usepackage{xcolor}
\usepackage{booktabs}
\usepackage{multirow}
\usepackage{sidecap}
\usepackage{amsthm}
\usepackage{amsmath}
\usepackage{amssymb}
\usepackage{amsfonts}
\usepackage{url}
\usepackage{pifont}
\usepackage{eqparbox}
\usepackage{arydshln} 
\usepackage{bigstrut}
\usepackage[linewidth=1pt]{mdframed}
\usepackage{framed}
\usepackage{float}
\restylefloat{table}

\newif\ifarxiv

\arxivtrue

\newcommand{\namecite}[1]{\citeauthor{#1}~\shortcite{#1}}
\newcommand\citet{\namecite}

\usepackage{color}
\definecolor{purple}{rgb}{0.5,0,1}
\definecolor{dcyan}{rgb}{0.2,0.6,0.5}
\definecolor{darkgreen}{rgb}{0,0.25,0}

\newcommand{\addedD}[1]{{\color{black} #1}}

\newcommand{\paran}[1]{\left( #1 \right)}
\newcommand{\brac}[1]{\left[ #1 \right]}
\newcommand{\braces}[1]{\left\lbrace #1 \right\rbrace}

\theoremstyle{definition}
\newtheorem{definition}{Definition}

\allowdisplaybreaks
\newcommand{\tableilp}{\textsc{TableILP}}
\newcommand{\paragram}{\textsc{Paragram}}

\newcommand{\tupleinf}{\textsc{TupleInf}}
\newcommand{\textilp}{\textsc{SemanticILP}}
\newcommand{\bidaf}{\textsc{BiDAF}}
\newcommand{\bidafTrained}{\textsc{BiDAF'}}
\newcommand{\proread}{\textsc{Proread}}
\newcommand{\syntprox}{\textsc{SyntProx}}

\newcommand{\cogcompnlp}{\textsc{CogCompNLP}}
\newcommand{\allennlp}{\textsc{AllenNLP}}

\newcommand\lucene{IR}

\newcommand{\processBank}{\textsc{ProcessBank}}
\newcommand{\regentsFourth}{\textsc{Regents 4th}}
\newcommand{\regentsEighth}{\textsc{Regents 8th}}
\newcommand{\publicFourth}{\textsc{AI2Public 4th}}
\newcommand{\publicEighth}{\textsc{AI2Public 8th}}

\newcommand{\spanLabelView}{\textsf{\small SpanLabelView}}
\newcommand{\sequence}{\textsf{\small Sequence}}
\newcommand{\spann}{\textsf{\small Span}}
\newcommand{\predArgView}{\textsf{\small PredArg}}
\newcommand{\semanticGraph}{\textsf{SemanticGraph}}
\newcommand{\clusterView}{\textsf{\small Cluster}}
\newcommand{\treeView}{\textsf{\small Tree}}
\newcommand{\tokens}{\textsf{\small Tokens}}
\newcommand{\shallowP}{\textsf{\small Shallow-Parse}}

\newcommand{\pos}{\textsf{\small POS}}
\newcommand{\lemmaa}{\textsf{\small Lemma}}
\newcommand{\quantities}{\textsf{\small Quantities}}
\newcommand{\coref}{\textsf{\small Coreference}}
\newcommand{\dep}{\textsf{\small Dependency}}
\newcommand{\ner}{\textsf{\small NER}}
\newcommand{\verbSRL}{\textsf{\small Verb-SRL}}
\newcommand{\prepSRL}{\textsf{\small Prep-SRL}}

\newcommand{\nomSRL}{\textsf{\small Nom-SRL}}
\newcommand{\commaSRL}{\textsf{\small Comma-SRL}}
\newcommand{\simple}{\textsf{\small Shallow}}
\newcommand{\ttitle}[1]{ \hspace{0.1cm} * \textit{#1}}
\newcommand{\know}[1]{\mathcal{K}(#1)}

\newcommand{\nodes}[1]{\mathbf{v}(#1)}
\newcommand{\edges}[1]{\mathbf{e}(#1)}
\newcommand{\score}[1]{\text{score}(#1)}
\newcommand{\eye}[1]{{\bf1}\!\!\braces{#1}}

\usepackage{array}
\newcolumntype{L}[1]{>{\raggedright\let\newline\\\arraybackslash\hspace{0pt}}m{#1}}
\newcolumntype{C}[1]{>{\centering\let\newline\\\arraybackslash\hspace{0pt}}m{#1}}
\newcolumntype{R}[1]{>{\raggedleft\let\newline\\\arraybackslash\hspace{0pt}}m{#1}}

\newcommand{\option}{a_{m}}

\newcommand{\setOf}[1]{\left\lbrace #1 \right\rbrace}
\newcommand{\onlyOne}{\exists !}

\newcommand{\specialcell}[2][c]{%
  \begin{tabular}[#1]{@{}c@{}}#2\end{tabular}}

\title{
\vspace*{-0.5in}
{{\small \hfill AAAI'18}\\
\vspace*{.25in}}
Question Answering as Global Reasoning over Semantic Abstractions}

\author{Daniel Khashabi
\thanks{Part of the work was done when the first and last authors were affiliated with the University of Illinois, Urbana-Champaign.}
\\
University of Pennsylvania \\
{\tt \small 
danielkh@cis.upenn.edu
}
    \\
    \And
  ~~~~Tushar Khot ~~~~ Ashish Sabharwal \\
  Allen Institute for Artificial Intelligence (AI2) \\
  {\tt \small tushark,ashishs@allenai.org}
  \And
  Dan Roth$^*$ \\
  University of Pennsylvania \\
  {\tt \small 
  danroth@cis.upenn.edu
  }
  }

\date{}

\begin{document}

\maketitle

\begin{abstract}
We propose a novel method for exploiting the semantic structure of text to answer multiple-choice questions. The approach is especially suitable for domains that require reasoning over a diverse set of linguistic constructs but have limited training data. To address these challenges, we present the first system, to the best of our knowledge, that reasons over a wide range of semantic abstractions of the text, which are derived using off-the-shelf, general-purpose, pre-trained natural language modules such as semantic role labelers, coreference resolvers, and dependency parsers. Representing multiple abstractions as a family of graphs, we translate question answering (QA) into a search for an optimal subgraph that satisfies certain global and local properties. This formulation generalizes several prior structured QA systems. Our system, \textilp, demonstrates strong performance on two domains simultaneously. In particular, on a collection of challenging science QA datasets, it outperforms various state-of-the-art approaches, including neural models, broad coverage information retrieval, and specialized techniques using structured knowledge bases, by 2\%-6\%. 
\end{abstract}

\section{Introduction}

Question answering (QA) is a fundamental challenge in AI, particularly in natural language processing (NLP) and reasoning. We consider the multiple-choice setting where $Q$ is a question, $A$ is a set of answer candidates, and the knowledge required for answering $Q$ is available in the form of raw text $P$. This setting applies naturally to reading comprehension questions, as well as to open format multiple-choice questions for which information retrieval (IR) techniques are able to obtain relevant text from a corpus.

Several state-of-the-art QA systems excel at locating the correct answer in $P$ based on proximity to question words, the distributional semantics of the text, answer type, etc \cite{aristo2016:combining}, however, often struggle with questions that require some form of reasoning or appeal to a more subtle understanding of the supporting text or the question. We demonstrate that we can use existing NLP modules, such as semantic role labeling (SRL) systems with respect to multiple predicate types (verbs, prepositions, nominals, etc.),
to derive multiple semantic views of the text and perform reasoning over these views to answer a variety of questions.

As an example, consider the following snippet of sports news text and an associated question:
\begin{framed}
\noindent\footnotesize\emph{\noindent$P$: Teams are under pressure after PSG purchased Neymar this season. \textbf{Chelsea purchased Morata}. The Spaniard looked like he was set for a move to Old Trafford for the majority of the summer only for Manchester United to sign Romelu Lukaku instead, paving the way for Morata to finally move to Chelsea for an initial \pounds 56m.
\\
\noindent
$Q$: Who did Chelsea purchase this season?  \\ 
$A$: \{\checkmark Alvaro Morata, Neymar, Romelu Lukaku \}
}
\end{framed}

Given the bold-faced text $P'$ in $P$, simple word-matching suffices to correctly answer $Q$. However, $P'$ could have stated the same information in many different ways. As paraphrases become more complex, they begin to involve more linguistic constructs such as coreference, punctuation, prepositions, and nominals. This makes understanding the text, and thus the QA task, more challenging.

For instead, $P'$ could instead say \emph{Morata is the recent \textbf{acquisition} by Chelsea}. This simple looking transformation can be surprisingly confusing for highly successful systems such as \bidaf~\cite{seo2016bidirectional}, which produces the partially correct phrase \emph{``Neymar this season.\ Morata''}. On the other hand, one can still answer the question confidently by abstracting relevant parts of $Q$ and $P$, and connecting them appropriately. Specifically, a verb SRL frame for $Q$ would indicate that we seek the object of the verb \emph{purchase}, a nominal SRL frame for $P'$ would capture that the \emph{acquisition} was of Morata and was done by Chelsea, and textual similarity would align \emph{purchase} with \emph{acquisition}.

\begin{figure}[ht]
    \centering
    \includegraphics[trim=-0.00cm 0cm 0cm -0.00cm, scale=0.55]{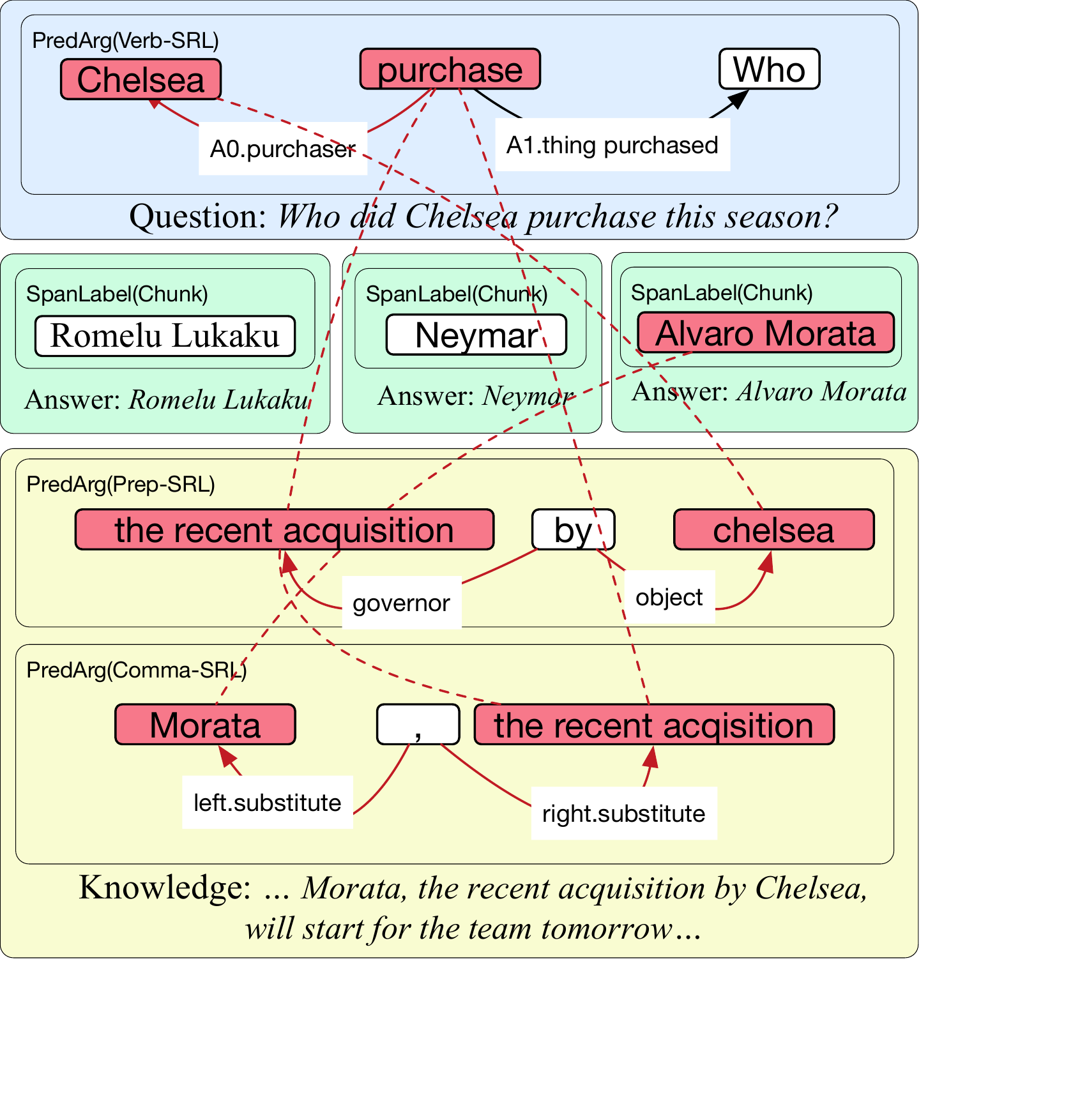}
    \caption{Depiction of \textilp\ reasoning for the example paragraph given in the text.
    Semantic abstractions of the question, answers, knowledge snippet are shown in different colored boxes (blue, green, and yellow, resp.). Red nodes and edges are the elements that are aligned (used) for supporting the correct answer. There are many other unaligned (unused) annotations associated with each piece of text that are omitted for clarity. 
    }
    \label{fig:morata-visualization}
\end{figure}

Similarly, suppose $P'$ instead said \emph{Morata\textbf{,} the recent acquisition \textbf{by} Chelsea\textbf{,} will start for the team tomorrow.} \bidaf\ now incorrectly chooses Neymar as the answer, presumably due to its proximity to the words \emph{purchased} and \emph{this season}. However, with the right abstractions, one could still arrive at the correct answer as depicted in Figure~\ref{fig:morata-visualization} for our proposed system, \textilp. This reasoning uses comma SRL to realize that the Morata is referring to the \emph{acquisition}, and a preposition SRL frame to capture that the acquisition was done \emph{by} Chelsea.

One can continue to make $P'$ more complex. For example, $P'$ could introduce the need for coreference resolution by phrasing the information as: \emph{\textbf{Chelsea} is hoping to have a great start this season by actively hunting for new players in the transfer period. Morata, the recent acquisition by the team, will start for \textbf{the team} tomorrow.} Nevertheless, with appropriate semantic abstractions of the text, the underlying reasoning remains relatively simple.

Given sufficiently large QA training data, one could conceivably perform end-to-end training (e.g., using a deep learning method) to address these linguistic challenges. However, existing large scale QA datasets such as SQuAD~\cite{Rajpurkar2016SQuAD10} often either have a limited linguistic richness or do not necessarily need reasoning to arrive at the answer~\cite{Jia2017AdversarialEF}. 
Consequently, the resulting models do not transfer easily to other domains.
For instance, the above mentioned BiDAF model trained on the SQuAD dataset performs substantially worse than a simple IR approach on our datasets. On the other hand, many of the QA collections in domains that require some form of reasoning, such as the science questions we use, are small  
(100s to 1000s of questions). This brings into question the viability of the aforementioned paradigm that attempts 
to learn everything from only the QA training data.

Towards the goal of effective structured reasoning in the presence of data sparsity, we propose to use a rich set of general-purpose, pre-trained NLP tools to create various \emph{semantic abstractions} of the raw text\footnote{This applies to all three inputs of the system: $Q$, $A$, and $P$.} in a domain independent fashion, as illustrated for an example in Figure~\ref{fig:morata-visualization}.
We represent these semantic abstractions as \emph{families of graphs}, where the family (e.g., trees, clusters, labeled predicate-argument graphs, etc.) is chosen to match the nature of the abstraction (e.g., parse tree, coreference sets, SRL frames, etc., respectively). The collection of semantic graphs is then augmented with inter- and intra-graph edges capturing lexical similarity (e.g., word-overlap score or word2vec distance).

This semantic graph representation allows us to formulate QA as the search for an optimal \emph{support graph}, a subgraph $G$ of the above augmented graph connecting (the semantic graphs of) $Q$ and $A$ via $P$. The reasoning used to answer the question is captured by a variety of requirements or constraints that $G$ must satisfy,
as well as a number of desired properties, encapsulating the ``correct'' reasoning, that makes $G$ preferable over other valid support graphs. For instance, a simple requirement is that $G$ must be connected and it must touch both $Q$ and $A$. Similarly, if $G$ includes a verb from an SRL frame, it is preferable to also include the corresponding subject. Finally, the resulting constrained optimization problem is formulated as an Integer Linear Program (ILP), and optimized using an off-the-shelf ILP solver.

This formalism may be viewed as a generalization of systems that reason over tables of knowledge~\cite{cohen2000:joins,tableilp2016:ijcai}: instead of operating over table rows (which are akin to labeled sequence graphs or predicate-argument graphs), we operate over a much richer class of semantic graphs. It can also be viewed as a generalization of the recent \tupleinf\ system~\cite{khot2017tupleinf}, which converts $P$ into a particular kind of semantic abstraction, namely Open IE tuples~\cite{Banko2007OpenIE}.

This generalization to multiple semantic abstractions poses two key technical challenges: (a) unlike clean knowledge-bases (e.g., \namecite{dong2015question}) used in many QA systems, abstractions generated from NLP tools (e.g., SRL) are noisy; and (b) even if perfect, using their output for QA requires delineating what information in $Q$, $A$, and $P$ is relevant for a given question, and what constitutes valid reasoning. The latter is especially challenging when combining information from diverse abstractions that, even though grounded in the same raw text, may not perfectly align. We address these challenges via our ILP formulation, by using our linguistic knowledge about the abstractions to design requirements and preferences for linking these abstractions.

We present a new QA system, \textilp,\footnote{Code available at: {https://github.com/allenai/semanticilp}} based on these ideas, and evaluate it on multiple-choice questions from two domains involving rich linguistic structure and reasoning: elementary and middle-school level science exams, and early-college level biology reading comprehension. Their data sparsity, as we show, limits the performance of state-of-the-art neural methods such as BiDAF~\cite{seo2016bidirectional}. \textilp, on the other hand, is able to successfully capitalize on existing general-purpose NLP tools in order to outperform existing baselines by 2\%-6\% on the science exams, leading to a new state of the art. It also generalizes well, as demonstrated by its strong performance on biology questions in the \processBank\ dataset~\cite{berant2014modeling}. Notably, while the best existing system for the latter relies on domain-specific structural annotation and question processing, \textilp\ needs neither.

\subsection{Related Work}
\label{subsec:related-work}

There are numerous QA systems operating over large knowledge-bases. Our work, however, is most closely related to systems that work either directly on the input text or on a semantic representation derived on-the-fly from it. In the former category are IR and statistical correlation based methods with surprisingly strong performance~\cite{aristo2016:combining}, as well as a number of recent neural architectures such as BiDAF~\cite{seo2016bidirectional}, Decomposable Attention Model~\cite{Parikh2016ADA}, etc. In the latter category are approaches that perform \emph{structured reasoning} over some abstraction
of text. For instance,
\addedD{\namecite{tableilp2016:ijcai} perform reasoning on the knowledge tables constructed using semi-automatical methods,} \namecite{khot2017tupleinf} use Open IE subject-verb-object relations~\cite{etzioni2008open}, \namecite{banarescu2013abstract} use AMR annotators \cite{wang2015transition}, and \namecite{krishnamurthy2016semantic} 
use a semantic parser \cite{zettlemoyer2012learning} to answer a given question. Our work differs in that we use of a wide variety of semantic abstractions simultaneously, and perform joint reasoning over them to identify which abstractions are relevant for a given question and how best to combine information from them.

\addedD{Our formalism can be seen as an extension of some of the prior work on structured reasoning over semi-structured text. For instance, in our formalism, each table used by \namecite{tableilp2016:ijcai} can be viewed as a semantic frame and represented as a predicate-argument graph. The table-chaining rules used there are equivalent to the reasoning we define when combining two annotation components. Similarly, Open IE tuples used by \cite{khot2017tupleinf} can also be viewed as a predicate-argument structure.}

One key abstraction we use is the predicate-argument structure provided by Semantic Role Labeling (SRL).
 Many SRL systems have been designed \cite{gildea2002automatic,PunyakanokRoYi08} using linguistic resources such as FrameNet~\cite{baker1998berkeley}, PropBank~\cite{kingsbury2002treebank}, and NomBank~\cite{meyers2004nombank}. These systems are meant to convey high-level information about predicates (which can be a verb, a noun, etc.) and related elements in the text. The meaning of each predicate is conveyed by a frame, the schematic representations of a situation. Phrases with similar semantics ideally map to the same frame and roles. Our system also uses other NLP modules, such as for coreference resolution\addedD{~\cite{lee2013deterministic}} and dependency parsing~\cite{chang2015illinoissl}.


While it appears simple to use SRL systems for QA~\cite{palmer2005proposition}, this has found limited success~\cite{kaisser2007question,pizzato2008indexing,Moreda2011CombiningSI}. The challenges earlier approaches faced were due to making use of VerbSRL only, while QA depends on richer information, not only verb predicates and their arguments, along with some level of brittleness of all NLP systems. \namecite{shen2007using} have partly addressed the latter challenge with an inference framework that formulates the task as a bipartite matching problem over the assignment of semantic roles, and managed to slightly improve QA. In this work we address both these challenges and go beyond the limitations of using a single predicate SRL system; we make use of SRL abstractions that are based on verbs, nominals, prepositions, and comma predicates, as well as textual similarity. We then develop an inference framework capable of exploiting combinations of these multiple SRL (and other) views, thus operating over a more complete semantic representation of the text.

A key aspect of QA is handling textual variations, on which there has been prior work using dependency-parse transformations~\cite{PunyakanokRoYi04a}. 
These approaches often define inference rules, which can generate new trees starting from a base tree. \namecite{bar2015knowledge} and \namecite{stern2012efficient} search over a space of a pre-defined set of text transformations (e.g., coreference substitutions, passive to active). Our work differs in that we consider a much wider range of textual variations by combining multiple abstractions, and make use of a more expressive inference framework.

\addedD{
Another aspect of our approach is the graphical representation of knowledge and the reasoning over it. We model the QA problem as a search in the space of potential support graphs. Search-based reasoning systems have been successful in several NLP areas~\cite{roth2004:ilp,chang2010discriminative,berant2010global,srikumar2011:srl,goldwasser2011:instructions,schuller2014tackling,fei2015profiler}. Compared to previous works, we use a larger set of annotators to account for diverse linguistic phenomena in questions and to handle errors of individual annotators. 
}

\section{Knowledge Abstraction and Representation}

We begin with our formalism for abstracting knowledge from text and representing it as a family of graphs, followed by specific instantiations of these abstractions using off-the-shelf NLP modules.

\subsection{Semantic Abstractions}
\label{subsec:kr}

\begin{figure}[ht]
\resizebox{\linewidth}{!}{
    \centering\includegraphics[trim=0cm 0cm 0cm 0.0cm,  scale=0.92]{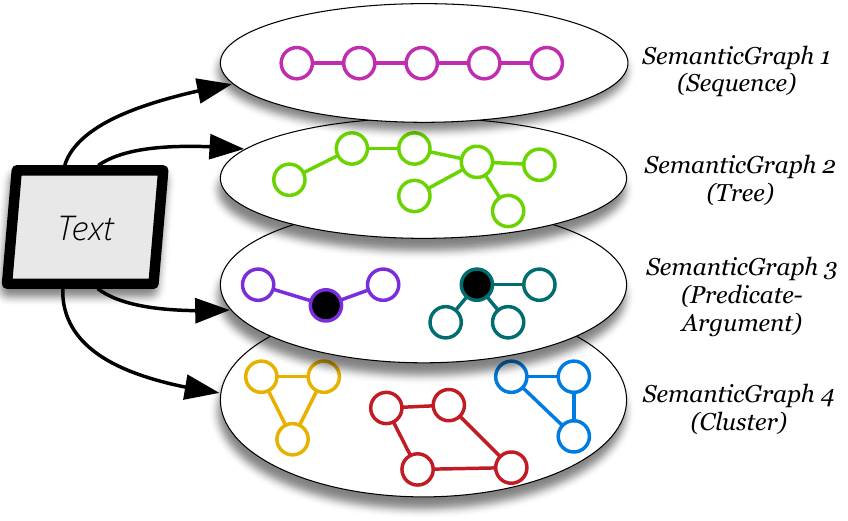}
    }
    \caption{Knowledge Representation used in our formulation. Raw text is associated with a collection of \semanticGraph s, which convey certain information about the text.
    There are implicit similarity edges among the nodes of the connected components of the graphs, and from nodes to the corresponding raw-text spans. }
    \label{fig:layers}
\end{figure}

The pivotal ingredient of the abstraction is raw text. This representation is used for question $Q$, each answer option $A_i$ and the knowledge snippet $P$, which potentially contains the answer to the question. The KB for a given raw text, consists of the text itself, embellished with various \semanticGraph s attached to it, as depicted in Figure~\ref{fig:layers}.  

Each \semanticGraph\ is representable from a family of graphs. In principle there need not be any constraints on the permitted graph families; however for ease of representation we choose the graphs to belong to one of the 5 following families:
\sequence\ graphs represent labels for each token in the sentence.
\spann\ family represents labels for spans of the text.  
\treeView, is a tree representation of text spans. 
\clusterView\ family, contain spans of text in different groups. 
\predArgView\ family represents predicates and their arguments; in this view edges represent the connections between each single predicates and its arguments. Each \semanticGraph\ belongs to one of the graph families and its content is determined by the semantics of the information it represents and the text itself.

We define the knowledge more formally here. For a given paragraph, $T$, its representation $\know{T}$ consists of a set of semantic graphs  $\know{T} = \setOf{g_1, g_2, \hdots}$. 
We define $\nodes{g} = \setOf{c_i}$ and $\edges{g} = \setOf{(c_{i}, c_{j})} $ to be the set of nodes and edges of a given graph, respectively. 

\subsection{Semantic Graph Generators}

Having introduced a graph-based abstraction for knowledge and categorized it into a family of graphs, we now delineate the instantiations we used for each family. Many of the pre-trained extraction tools we use are available in \cogcompnlp.\footnote{Available at: http://github.com/CogComp/cogcomp-nlp}

\begin{itemize}[leftmargin=0.25cm]
    \item 
         \sequence\, or labels for sequence of tokens; for example \lemmaa\ and \pos\  \cite{RothZe98}. 
    \item
        \spann\ which can contains labels for spans of text; we instantiated \shallowP\ \cite{PunyakanokRo01}, \quantities\ \cite{RoyViRo15}, \ner\  \cite{RatinovRo09,RedmanSaRo17}).   
    \item 
        \treeView, a tree representation connecting spans of text as nodes; for this we used \dep\ of \namecite{chang2015illinoissl}. 
    \item 
        \clusterView, or spans of text clustered in groups. An example is \coref\ \cite{lee2011stanford}. 
    \item 
        \predArgView; for this view we used \verbSRL\ and \nomSRL \cite{PunyakanokRoYi08,roth2016neural}, \prepSRL\ \cite{SrikumarRo13}, \commaSRL\ \cite{arivazhagan2016labeling}.
\end{itemize}

Given \semanticGraph\ generators we have the question, answers and paragraph represented as a collection of graphs. Given the instance graphs, creating augmented graph will be done implicitly as an optimization problem in the next step.

\section{QA as Reasoning Over Semantic Graphs}

We introduce our treatment of QA as an optimal subgraph selection problem over knowledge. 
We treat question answering as the task of finding the best support in the knowledge snippet, for a given question and answer pair,
measured in terms of the strength of a ``support graph'' defined as follows.

The inputs to the QA system are, a question $\know{Q}$, the set of answers $\setOf{\know{A_i}}$ and given a knowledge snippet $\know{P}$.\footnote{For simplicity, from now on, we drop ``knowledge"; e.g., instead of saying ``question knowledge", we say ``question". } Given such representations, we will form a reasoning problem, which is formulated as an optimization problem, searching for a ``support graph" that connects the question nodes to a unique answer option through nodes and edges of a snippet. 

Define the instance graph $I = I(Q, \setOf{A_i}, P)$ as the union of knowledge graphs: $I \triangleq \know{Q} \cup \paran{\know{A_i}} \cup \know{P}$. Intuitively, we would like the support graph to be connected, and to include nodes from the question, the answer option, and the knowledge. Since the \semanticGraph\ is composed of many disjoint sub-graph, we define \emph{augmented graph} $I^+$ to 
model a bigger structure over the instance graphs $I$. Essentially we \textit{augment} the instance graph and weight the new edges. Define a scoring function $f:(v_1, v_2) $ labels pair of nodes $v_1$ and $v_2$ with an score which represents their phrase-level entailment or similarity. 




\begin{definition}
\label{def:augmented-graph}
An \emph{augmented graph} $I^+$, for a question $Q$, answers $\setOf{A_i}$ and knowledge $P$, is defined with the following properties: 
\begin{enumerate}
    \item Nodes: $\nodes{I^+} = \nodes{I(Q, \setOf{A_i}, P)}$
    \item Edges:\footnote{Define $\know{T_1} \otimes \know{T_2} \triangleq \bigcup_{\substack{ (g_1, g_2) \in \\ \know{T_1} \times \know{T_2}} } \nodes{g_1} \times \nodes{g_2}$, where $\nodes{g_1} \times \nodes{g_2} = \setOf{ (v, w); v \in \nodes{g_1}, w \in \nodes{g_2} }. $} 
    \begin{align*} 
     \edges{I^+} &= \edges{I} \cup  \know{Q} \otimes \know{P} \cup \brac{\cup_i \know{P} \otimes \know{A_i} }
    \end{align*} 
    \item Edge weights: for any $e \in I^+$: 
    \begin{itemize}
      \item If $e \notin I$, the edge connects two nodes in different connected components: 
      $$
      \forall e = (v_1, v_2) \notin I : w(e) = f(v_1, v_2) 
      $$
      \item If $e \in I$, the edge belongs to a connected component, and the edge weight information about the reliability of the \semanticGraph\ and semantics of the two nodes. 
      $$
        \forall g \in I, \forall e \in g: w(e) = f'(e, g)
      $$
    \end{itemize}
\end{enumerate}
\end{definition}

Next, we have to define \textit{support graphs}, the set of graphs that support the reasoning of a question. For this we will apply some structured constraints on the augmented graph. 

\begin{definition}
\label{def:support-graph}
A \emph{support graph} $G = G(Q,\setOf{A_i}, P)$ for a question $Q$, answer-options $\setOf{A_i}$ and paragraph $P$, is a subgraph $(V,E)$ of $I^+$ with the following properties:
\begin{enumerate}
\item $G$ is connected.
\item $G$ has intersection with the question, the knowledge, and exactly one answer candidate:\footnote{$\onlyOne$ here denotes the uniqueness quantifier, meaning ``there exists one and only one".} 
$$
G \cap \know{Q} \neq \emptyset,
\hspace{0.3cm}  G \cap \know{P} \neq \emptyset,
\hspace{0.3cm}  \onlyOne \; i: G \cap \know{A_i} \neq \emptyset 
$$
\item $G$ satisfies structural properties per each connected component, as summarized in Table~\ref{table:structural-properties}. 
\end{enumerate}
\end{definition}

\begin{table}
    \centering
    \small 
    \setlength\tabcolsep{10pt}
    \setlength\doublerulesep{\arrayrulewidth}
    \resizebox{\linewidth}{!}{
    \begin{tabular}{C{1.9cm}|L{5.1cm}}
        Sem. Graph &  Property \\
        \hline\hline \bigstrut[t]
         \predArgView & Use at least (a) a predicate and its argument, or (b) two arguments  \\ 
         \hline 
         \clusterView & Use at least two nodes  \\
         \hline 
         \treeView & Use two nodes with distance less than $k$ \\ 
         \hline 
         \spanLabelView & Use at least $k$ nodes   
    \end{tabular}
    }
    \caption{Minimum requirements for using each family of graphs. Each graph connected component (e.g. a \predArgView\ frame, or a \coref\ chain) cannot be used unless the above-mentioned conditioned is satisfied. }
    \label{table:structural-properties}
\end{table}

Definition~\ref{def:support-graph} characterizes what we call a potential solution to a question. A given question and paragraph give rise to a large number of possible support graphs. We define the space of feasible support graphs as $\mathcal{G}$ (i.e., all the graphs that satisfy Definition~\ref{def:support-graph}, for a given $(Q, \setOf{A_i}, P)$).
To rank various feasible support graphs in such a large space, we define a scoring function $\score{G}$ as:
\begin{equation}
\label{eq:score-function}
\sum_{v \in \nodes{G}} w(v) + \sum_{e \in \edges{G}} w(e) - \sum_{c \in \mathcal{C} } w_c \, \eye{c \text{ is \addedD{violated}}}
\end{equation}
for some set of preferences (or soft-constraints) $\mathcal{C}$.
\addedD{When $c$ is violated, denoted by the indicator function $\eye{c \text{ is violated}}$ in Eq.~(\ref{eq:score-function}), we penalize the objective value by some fixed amount $w_c$.}
The second term is supposed to bring more sparsity to the desired solutions, just like how regularization terms act in machine learning models \cite{natarajan1995sparse}. The first term is the sum of weights we defined when constructing the augmented-graph, and is supposed to give more weight to the models that have better and more reliable alignments between its nodes. 
The role of the inference process will be to choose the ``best" one under our notion of \emph{desirable} support graphs: 
\begin{equation}
\label{eq:objective}
G^* = \argmax_{G \in \mathcal{G}} \score{G}
\end{equation}

\subsection{ILP Formulation}
\label{subsec:ilp}

Our QA system, \textilp, models the above support graph search of Eq.~(\ref{eq:objective}) as an ILP optimization problem, i.e., as maximizing a linear objective function over a finite set of variables, subject to a set of linear inequality constraints. A summary of the model is given below. 

The augmented graph is not explicitly created; instead, it is implicitly created. The nodes and edges of the augmented graph are encoded as a set of binary variables. The value of the binary variables reflects whether a node or an edge is used in the optimal graph $G^*$. The properties listed in Table~\ref{table:structural-properties} are implemented as weighted linear constraints using the variables defined for the nodes and edges. 

As mentioned, edge weights in the augmented graph come from a function, $f$, which captures (soft) phrasal entailment between question and paragraph nodes, or paragraph and answer nodes, to account for lexical variability.
\addedD{
In our evaluations, we use two types of $f$. (a) Similar to \namecite{tableilp2016:ijcai}, we use a WordNet-based \cite{miller1995wordnet} function to score word-to-word alignments, and use this as a building block to compute a phrase-level alignment score as the weighted sum of word-level alignment scores.
Word-level scores are computed using WordNet's hypernym and synonym relations, and weighted using relevant word-sense frequency.
$f$ for similarity (as opposed to entailment) is taken to be the average of the entailment scores in both directions. (b) For longer phrasal alignments (e.g., when aligning phrasal verbs) we use the Paragram system of \namecite{wieting2015paraphrase}.
}

The final optimization is done on Eq.~(\ref{eq:score-function}). The first part of the objective is the sum of the weights of the sub-graph, which is what an ILP does, since the nodes and edges are modeled as variables in the ILP. The second part of Eq.~(\ref{eq:score-function}) contains a set of preferences $\mathcal{C}$, summarized in Table~\ref{tab:preferences}, meant to apply \textit{soft} structural properties that partly dependant on the knowledge instantiation. These preferences are soft in the sense that they are applied with a weight to the overall scoring function (as compare to a hard constraint). For each preference function $c$  there is an associated binary or integer variable with weight $w_c$, and we create appropriate constraints to simulate the corresponding behavior. 

\begin{table}
    \begin{framed}
    \noindent - Number of sentences used is more than $k$\\ 
    - Active edges connected to each chunk of the answer option, more than $k$ \\ 
    - More than k chunks in the active answer-option \\ 
    - More than k edges to each question constituent  \\ 
    - Number of active question-terms  \\ 
    - If using \predArgView of $\know{Q}$, at least an argument should be used  \\ 
    - If using \predArgView(\verbSRL) of $\know{Q}$, at least one predicate should be used.
    \end{framed}
    \caption{The set of preferences functions in the objective. }
    \label{tab:preferences}
\end{table}

We note that the ILP objective and constraints aren't tied to the particular domain of evaluation; they represent general properties that capture what constitutes a well supported answer for a given question.

\section{Empirical Evaluation}
\label{sec:experiments}

We evaluate on two domains that differ in the nature of the supporting text (concatenated individual sentences vs.\ a coherent paragraph), the underlying reasoning, and the way questions are framed. We show that \textilp\ outperforms a variety of baselines, including retrieval-based methods, neural-networks, structured systems, and the current best system for each domain. These datasets and systems are described next, followed by results.

\subsection{Question Sets}

For the first domain, we have a collection of question sets containing elementary-level science questions from standardized tests \cite{aristo2016:combining,khot2017tupleinf}. Specifically, \regentsFourth\ contains all non-diagram multiple choice questions from 6 years of NY Regents 4th grade science exams (127 train questions, 129 test). \regentsEighth\ similarly contains 8th grade questions (147 train, 144 test). The corresponding expanded datasets are \publicFourth\ (432 train, 339 test) and \publicEighth\ (293 train, 282 test).\footnote{AI2 Science Questions V1 at http://data.allenai.org/ai2-science-questions}

For the second domain, we use the \processBank\footnote{https://nlp.stanford.edu/software/bioprocess} dataset for the reading comprehension task proposed by \namecite{berant2014modeling}. It contains paragraphs about biological processes and two-way multiple choice questions about them. We used a broad subset of this dataset that asks about events or about an argument that depends on another event or argument.\footnote{These are referred to as ``dependency questions" by \namecite{berant2014modeling}, and cover around 70\% of all questions.}. The resulting dataset has 293 train and 109 test questions, based on 147 biology paragraphs.

Test scores are reported as percentages. For each question, a system gets a score of $1$ if it chooses the correct answer, $1/k$ if it reports a $k$-way tie that includes the correct answer, and $0$ otherwise.

\subsection{Question Answering Systems}
\label{section:baslines}
We consider a variety of baselines, including the best system for each domain.

\textbf{\lucene} (information retrieval baseline).
We use the IR solver from \namecite{aristo2016:combining}, which selects the answer option that has the best matching sentence in a corpus. The sentence is forced to have a non-stopword overlap with both $q$ and $a$.

\textbf{\textilp} (our approach). 
Given the input instance (question, answer options, and a paragraph), we invoke various NLP modules to extract semantic graphs. We then generate an ILP and solve it using the open source SCIP engine~\cite{solver:scip}, returning the active answer option $\option$ from the optimal solution found. To check for ties, we disable $\option$, re-solve the ILP, and compare the score of the second-best answer, if any, with that of the best score.

For the science question sets, where we don't have any paragraphs attached to each question, we create a passage by using the above \lucene\ solver to retrieve scored sentences for each answer option and then combining the top 8 unique sentences (across all answer options) to form a paragraph.

\addedD{
While the sub-graph optimization can be done over the entire augmented graph in one shot, 
our current implementation uses multiple simplified solvers, each performing reasoning over augmented graphs for a commonly occurring annotator combination, as listed in Table~\ref{tab:combinations}. For all of these annotator combinations, we let the representation of the answers be $\know{A}=\{\shallowP, \tokens\}$. Importantly, our choice of working with a few annotator combinations is mainly for simplicity of implementation and suffices to demonstrate that reasoning over even just two annotators at a time can be surprisingly powerful. 
There is no fundamental limitation in implementing \textilp\ using one single optimization problem as stated in Eq.~(\ref{eq:objective}). 

\begin{table}
    \small 
  \setlength\tabcolsep{2pt}
  \setlength\doublerulesep{\arrayrulewidth}
    \centering
    \resizebox{\linewidth}{!}{
    \begin{tabular}{c|l}
        Combination & \multicolumn{1}{c}{Representation}   \\
        \hline\hline \bigstrut[t]
         \multirow{2}{*}{
            Comb-1
         } 
         
         & $\know{Q}=\{\shallowP, \tokens\}$  \\ 
           &  $\know{P}=\{\shallowP, \tokens, \dep\}$  \\ 
           \hline 
         \multirow{2}{*}{
            Comb-2
         } &$\know{Q}=\setOf{\verbSRL, \shallowP}$    \\ 
          & $\know{P}=\setOf{ \verbSRL }$ \\ 
         \hline 
          \multirow{2}{*}{
            Comb-3 
          } & $\know{Q}=\setOf{\verbSRL, \shallowP}$    \\  
          &   $\know{P}=\{\verbSRL, \coref \}$  \\  
         \hline 
         \multirow{2}{*}{
            Comb-4
         } & $\know{Q}=\setOf{ \verbSRL,\shallowP}$    \\ 
          &  $\know{P}=\setOf{\commaSRL }$ \\ 
         \hline 
         \multirow{2}{*}{
            Comb-5 
         } & $\know{Q}=\setOf{\verbSRL,\shallowP }$  \\ 
          & $\know{P}=\setOf{ \prepSRL }$ \\ 
    \end{tabular}
    }
    \caption{The semantic annotator combinations used in our implementation of \textilp. 
    }
    \label{tab:combinations}
\end{table}

Each simplified solver associated with an annotator combination in Table~\ref{tab:combinations} produces a confidence score for each answer option. We create an \emph{ensemble} of these solvers as a linear combination of these scores, with weights trained using the union of training data from all questions sets. 
}

\textbf{\bidaf} (neural network baseline).
We use the recent deep learning reading comprehension model of \namecite{seo2016bidirectional}, which is one of the top performing systems on the SQuAD dataset and has been shown to generalize to another domain as well~\cite{Min2017QuestionAT}. Since \bidaf\ was designed for fill-in-the-blank style questions, we follow the variation used by \namecite{Kembhavi2017AreYS} to apply it to our multiple-choice setting. Specifically, we compare the predicted answer span to each answer candidate and report the one with the highest similarity.

We use two variants: the original system, \bidaf, pre-trained on 100,000+ SQuAD questions, as well as an extended version, \bidafTrained, obtained by performing continuous training to fine-tune the SQuAD-trained parameters using our (smaller) training sets. For the latter, we convert multiple-choice questions into reading comprehension questions by generating all possible text-spans within sentences, with token-length at most \textit{correct answer length + 2}, and choose the ones with the highest similarity score with the correct answer. We use the \allennlp\ re-implementation of \bidaf\footnote{Available at: https://github.com/allenai/allennlp}, train it on SQuAD, followed  by training it on our dataset. We tried different variations (epochs and learning rates) and selected the model which gives the best average score across all the datasets. As we will see, the variant that was further trained on our data often gives better results. 

\textbf{\tupleinf} (semi-structured inference baseline).
Recently proposed by \namecite{khot2017tupleinf}, this is a state-of-the-art system designed for science questions. It uses Open IE~\cite{Banko2007OpenIE} tuples derived from the text as the knowledge representation, and performs reasoning over it via an ILP. It has access to a large knowledge base of Open IE tuples, and exploits redundancy to overcome challenges introduced by noise and linguistic variability.

\textbf{\proread} and \textbf{\syntprox}. 
\proread is a specialized and best performing system on the \processBank\ question set. \namecite{berant2014modeling} annotated the training data with events and event relations, and trained a system to extract the process structure. Given a question, \proread\ converts it into a query (using regular expression patterns and keywords) and executes it on the process structure as the knowledge base. Its reliance on a question-dependent query generator and on a process structure extractor makes it difficult to apply to other domains.

\syntprox\ is another solver suggested by \cite{berant2014modeling}. It aligns content word lemmas in both the question and the answer against the paragraph, and select the answer tokens that are closer to the aligned tokens of the questions. The distance is measured using dependency tree edges. To support multiple sentences they connect roots of adjacent sentences with bidirectional edges.

\subsection{Experimental Results}
We evaluate various QA systems on datasets from the two domains. The results are summarized below, followed by some some insights into \textilp's behavior and an error analysis. 
\paragraph{Science Exams.}
The results of experimenting on different grades' science exams are summarized in Table~\ref{table:science}, which shows the exam scores as a percentage.  
The table demonstrates that \textilp\ consistently outperforms the best baselines in each case by 2\%-6\%. 

Further, there is no absolute winner among the baselines; while \lucene\ is good on the 8th grade questions, \tupleinf\ and \bidafTrained\ are better on 4th grade questions. This highlights the differing nature of questions for different grades. 
\begin{table}[htb]
  \centering
  \small
  \setlength\tabcolsep{2pt}
  \setlength\doublerulesep{\arrayrulewidth}
  \resizebox{\linewidth}{!}{
  \begin{tabular}{l|cccc|c}
  Dataset & \bidaf & \bidafTrained & \lucene & \tupleinf & \textilp \\ 
  \hline\hline \bigstrut[t]
  \regentsFourth  & 56.3 & 53.1 & 59.3 & \emph{61.4} & {\bf 67.6} \\ 
  \publicFourth   & 50.7 & \emph{57.4} & 54.9 & 56.1 & {\bf 59.7} \\ 
  \regentsEighth  & 53.5 & 62.8 & \emph{64.2} & 61.3 & {\bf 66.0} \\ 
  \publicEighth   & 47.7 & 51.9 & \emph{52.8} & 51.6 & {\bf 55.9} \\ 
  \hline
  \end{tabular}
  }
  \caption{Science test scores as a percentage. On elementary level science exams, \textilp\ consistently outperforms baselines. In each row, the best score is in \textbf{bold} and the best baseline is \emph{italicized}.  }
  \label{table:science}
\end{table}

\paragraph{Biology Exam.} 
The results on the \processBank\ dataset are summarized in Table~\ref{table:bio}. While \textilp's performance is substantially better than most baselines and close to that of \proread, it is important to note that this latter baseline enjoys additional supervision of domain-specific event annotations. This, unlike our other relatively general baselines, makes it limited to this dataset,
which is also why we don't include it in Table~\ref{table:science}.

We evaluate \lucene\ on this reading comprehension dataset by creating an ElasticSearch index, containing the sentences of the knowledge paragraphs. 

\begin{table}[htb]
  \centering
  \small
  \setlength\tabcolsep{2pt}
  \setlength\doublerulesep{\arrayrulewidth}
\resizebox{\linewidth}{!}{
  \begin{tabular}{c|cccc|c}
  \proread &  \syntprox & \lucene &  \bidaf & \bidafTrained & \textilp \\ 
  \hline\hline \bigstrut[t]
  68.1  &  61.9 & 63.8 & 58.7 & 61.3 & {\bf 67.9 }\\ 
  \hline
  \end{tabular}
  }
  \caption{Biology test scores as a percentage. \textilp\ outperforms various baselines on the \processBank\ dataset and roughly matches the specialized best method.}
  \label{table:bio}
\end{table}

\subsection{Error and Timing Analysis}
For some insight into the results, we include a brief analysis of our system's output compared to that of other systems.

We identify a few main reasons for \textilp's errors.
Not surprisingly, some mistakes
(see Appendix figure for an example)
 can be traced back to failures in generating proper annotation (\semanticGraph). Improvement in SRL modules or redundancy can help address this. 
Some mistakes are from the current ILP model not supporting the ideal reasoning, i.e., the requisite knowledge exists in the annotations, but the reasoning fails to exploit it. 
Another group of mistakes is due to the complexity of the sentences, and the system lacking a way to represent the underlying phenomena with our current annotators. 

A weakness (that doesn't seem to be particular to our solver) is reliance on explicit mentions. If there is a meaning indirectly implied by the context and our annotators are not able to capture it, our solver will miss such questions. There will be more room for improvement on such questions with the development of discourse analysis systems.   

When solving the questions that don't have an attached paragraph, relevant sentences need to be fetched from a corpus. A subset of mistakes on this dataset occurs because the extracted knowledge does not contain the correct answer.

\subsubsection{ILP Solution Properties.}


Our system is implemented using many constraints, requires using many linear inequalities which get instantiated on each input instanced, hence there are a different number of variables and inequalities for each input instance. There is an overhead time for pre-processing an input instance, and convert it into an instance graph. Here in the timing analysis we provide we ignore the annotation time, as it is done by black-boxes outside our solver.

Table~\ref{tab:stats} summarizes various ILP and support graph statistics for \textilp, averaged across \processBank\ questions. Next to \textilp\ we have included numbers from \tableilp\ which has similar implementation machinery, but on a very different representation. While the size of the model is a function of the input instance, on average, \textilp\ tends to have a bigger model (number of constraints and variables). The model creation time is significantly time-consuming in \textilp\ as involves many graph traversal operations and jumps between nodes and edges. 
We also providing times statistics for \tupleinf\, which takes roughly half the time of \tableilp, which means that it is faster than \textilp.

\begin{table}[htb]
\centering
\footnotesize
\setlength\tabcolsep{5pt}
\setlength\doublerulesep{\arrayrulewidth}
\resizebox{\linewidth}{!}{
\begin{tabular}{llccc}
\multirow{2}{*}{Category} & 
\multirow{2}{*}{Quantity} & 
Avg. & Avg. & Avg.  \\
& & { \tiny (\textilp) } & { \tiny (\tableilp) } & { \tiny (\tupleinf) } \\ 
\hline\hline \bigstrut[t]
\multirow{3}{*}{ILP complexity} & \#variables & 2254.9 & 1043.8 & 1691.0 \\
& \#constraints & 4518.7 & 4417.8 & 4440.0 \\
\hline \bigstrut[t]
\multirow{2}{*}{Timing stats} & model creation & 5.3 sec  & 1.9 sec & 1.7 sec \\
& solving the ILP & 1.8 sec   & 2.1 sec &  0.3 sec\\
\hline
\end{tabular}}
\caption{\textilp\ statistics averaged across questions, as compared to  \tableilp\ and \tupleinf\ statistics.
}
\label{tab:stats}
\end{table}

\subsection{Ablation Study}
In order to better understand the results, we ablate the contribution of different annotation combinations, where we drop different combination from the ensemble model. We retrain the ensemble, after dropping each combination.

The results are summarized in Table~\ref{tab:ablation}. While Comb-1
seems to be important for science tests, it has limited contribution to the biology tests. On 8th grade exams, the \verbSRL\ and \commaSRL-based alignments provide high value. Structured combinations (e.g., \verbSRL-based alignments) are generally more important for the biology domain. 

\begin{table}[htb]
\centering
\footnotesize
\setlength\tabcolsep{5pt}
\setlength\doublerulesep{\arrayrulewidth}
\resizebox{\linewidth}{!}{
\begin{tabular}{lC{2.2cm}C{2.2cm}}
 & \publicEighth & \processBank  \\ 
\hline\hline
Full \textilp\ & 
55.9 &	67.9 \\ 
\hline 
no Comb-1  
&
 -3.1 &	-1.8 \\ 
no Comb-2 
& -2.0 &	-4.6 \\ 
no Comb-3  
& -0.6 &	-1.8 \\ 
no Comb-4 
& -3.1 &	-1.8 \\ 
no Comb-5  
& -0.1 &	-5.1 \\ 
\hline
\end{tabular}
}
\caption{Ablation study of \textilp\ components on various datasets. The first row shows the overall test score of the full system, while other rows report the change in the score as a result of dropping an individual combination. The combinations are listed in Table~\ref{tab:combinations}.}
\label{tab:ablation}
\end{table}

\begin{figure}[ht]
    \centering
    \includegraphics[scale=0.75]{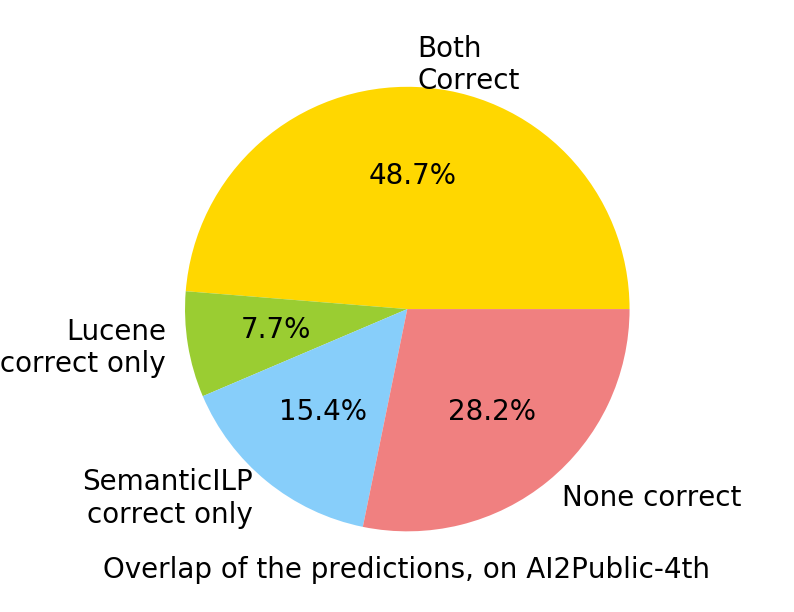}
    \caption{Overlap of the predictions of \textilp\ and \lucene\ on 50 randomly-chosen questions from \publicFourth.
    }
    \label{fig:piechart}
\end{figure}

\paragraph{Complementarity to \lucene.}
Given that in the science domain the input snippets fed to \textilp\ are retrieved through a process similar to the \lucene\ solver, one might naturally expect some similarity in the predictions. 
The pie-chart in Figure~\ref{fig:piechart} shows the overlap between mistakes and correct predictions of \textilp\ and \lucene\ on 50 randomly chosen training questions from \publicFourth. While there is substantial overlap in questions that both answer correctly (the yellow slice) and both miss (the red slice), there is also a significant number of questions solved by \textilp\ but not \lucene\ (the blue slice), almost twice as much as the questions solved by \lucene but not \textilp\ (the green slice).  

\begin{figure*}
    \centering
    \hspace{0.8cm}
    \includegraphics[scale=0.39]{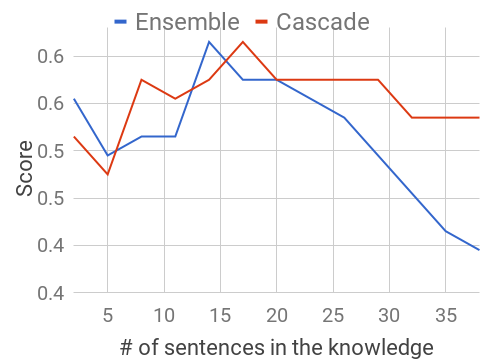}
    \hspace{0.8cm}
    \includegraphics[scale=0.39]{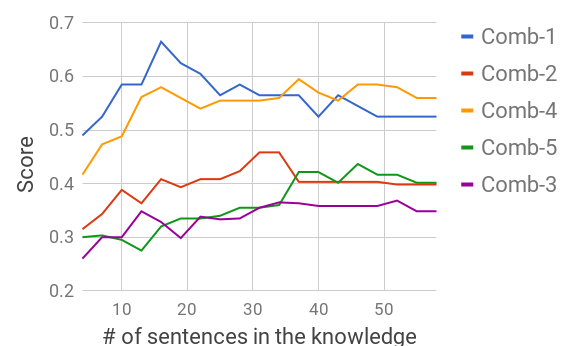}
    \caption{Performance change for varying knowledge length.}
    \label{fig:knowledge-experiment}
\end{figure*}

\subsubsection{Cascade Solvers.}
In Tables~\ref{table:science} and~\ref{table:bio}, we presented one \textit{single} instance of \textilp\ with state-of-art results on multiple datasets, where the solver was an ensemble of semantic combinations (presented in Table~\ref{tab:combinations}). Here we show a simpler approach that achieves stronger results on individual datasets, at the cost of losing a little generalization across domains. 
Specifically, we create two ``cascades'' (i.e., decision lists) of combinations, where the ordering of combinations in the cascade is determined by the training set precision of the simplified solver representing an annotator combination (combinations with higher precision appear earlier). One cascade solver targets science exams and the other the biology dataset.

The results are reported in Table~\ref{table:cascade}. On the 8th grade data, the cascade solver created for science test achieves higher scores than the generic ensemble solver. Similarly, the cascade solver on the biology domain outperforms the ensemble solver on the \processBank\ dataset.

\begin{table}[htb]
  \centering
  \small
  \setlength\tabcolsep{5pt}
  \setlength\doublerulesep{\arrayrulewidth}
\resizebox{\linewidth}{!}{
  \begin{tabular}{cl|ccc}
  \multicolumn{2}{c|}{Dataset} & Ensemble & \specialcell{Cascade \\(Science)} & \specialcell{Cascade \\(Biology)} \\ 
  \hline\hline \bigstrut[t]
   \parbox[t]{2ex}{\multirow{4}{*}{\rotatebox[origin=c]{90}{Science}}} 
   &  \regentsFourth  & {\bf 67.6} & 64.7 & 63.1 \\ 
   & \publicFourth   & {\bf 59.7} & 56.7 & 55.7\\ 
   & \regentsEighth  & 66.0 & {\bf 69.4} & 60.3 \\ 
   & \publicEighth   & 55.9 & {\bf 56.5 } & 54.3 \\ 
  \hline 
    & \processBank   & 67.9 & 59.6 & {\bf 68.8 } \\ 
  \hline
  \end{tabular}
  }
  \caption{
    Comparison of test scores of \textilp\ using a generic ensemble vs.\ domain-targeted cascades of annotation combinations. 
  }
  \label{table:cascade}
\end{table}

\paragraph{Effect of Varying Knowledge Length.}
We analyze the performance of the system as a function of the length of the paragraph fed into \textilp, for 50 randomly selected training questions from the \regentsFourth\ set. Figure~\ref{fig:knowledge-experiment} (left) shows the overall system, for two combinations introduced earlier, as a function of knowledge length, counted as the number of sentences in the paragraph.

As expected, the solver improves with more sentences, until around 12-15 sentences, after which it starts to worsen with the addition of more irrelevant knowledge. While the cascade combinations did not show much generalization across domains, they have the advantage of a smaller drop when adding irrelevant knowledge compared to the ensemble solver. This can be explained by the simplicity of cascading and minimal training compared to the ensemble of annotation combinations. 

Figure~\ref{fig:knowledge-experiment} (right) shows the performance of individual combinations as a function of knowledge length.
It is worth highlighting that while Comb-1
(blue) often achieves higher coverage and good scores in simple paragraphs (e.g., science exams), it is highly sensitive to knowledge length. On the other hand,  highly-constrained combinations have a more consistent performance with increasing knowledge length, at the cost of lower coverage.

\section{Conclusion}
\label{sec:conclusion}

Question answering is challenging in the presence of linguistic richness and diversity, especially when arriving at the correct answer requires some level of reasoning. Departing from the currently popular paradigm of generating a very large dataset and learning ``everything'' from it in an end-to-end fashion, we argue---and demonstrate via our QA system---that one can successfully leverage pre-trained NLP modules to extract a sufficiently complete linguistic abstraction of the text that allows answering interesting questions about it. This approach is particularly valuable in settings where there is a small amount of data. Instead of exploiting peculiarities of a large but homogeneous dataset, as many state-of-the-art QA systems end up doing, we focus on confidently performing certain kinds of reasoning, as captured by our semantic graphs and the ILP formulation of support graph search over them. Bringing these ideas to practice, our system, \textilp, achieves state-of-the-art performance on two domains with very different characteristics, outperforming both traditional and neural models.

\subsection*{Acknowledgments}
The authors would like to thank
Peter Clark, 
Bhavana Dalvi,
Minjoon Seo, 
Oyvind Tafjord,  
Niket Tandon,
and  
Luke Zettlemoyer
for valuable discussions and help.

This work is supported by a gift from AI2 and by contract FA8750-13-2-0008 with the US Defense Advanced Research Projects Agency (DARPA). The views expressed are those of the authors and do not reflect the official policy or position of the U.S. Government.

\bibliographystyle{aaai}
\begin{small}
\bibliography{extrabib} 
\end{small}

\ifarxiv 
\input{appendix}

\fi

\input{appendix}
\end{document}

%% file: appendix.tex
\clearpage

\appendix

\section{Appendix: Reasoning Formulation}

Here we provide some details of our reasoning formulation and its implementation as an ILP.

The support graph search for QA is modeled as an ILP optimization problem, i.e., as maximizing a linear objective function over a finite set of variables, subject to a set of linear inequality constraints.
We note that the ILP objective and constraints aren't tied to the particular domain of evaluation; they represent general properties that capture what constitutes a well-supported answer for a given question.

Our formulation consists of multiple kinds of reasoning, encapsulating our semantic understanding of the types of knowledge (e.g., verbs, corefs, etc.) extracted from the text, the family to graphs used to represent them, and how they interact in order to provide support for an answer candidate. Each kind of reasoning is added to a general body, defined in Table~\ref{tab:general-body}, that is shared among different reasoning types. This general body encapsulates basic requirements such as at most one answer candidate being chosen in the resulting support graph.

In what follows we delineate various forms of reasoning, capturing different semantic abstractions and valid interactions between them.

\subsubsection{Comb-1 (\simple\ Alignment)}
\begin{figure*}[th]
    \centering
    \fbox{\includegraphics[trim=1.1cm 0.6cm 4.8cm .7cm, width=490pt,clip=true]{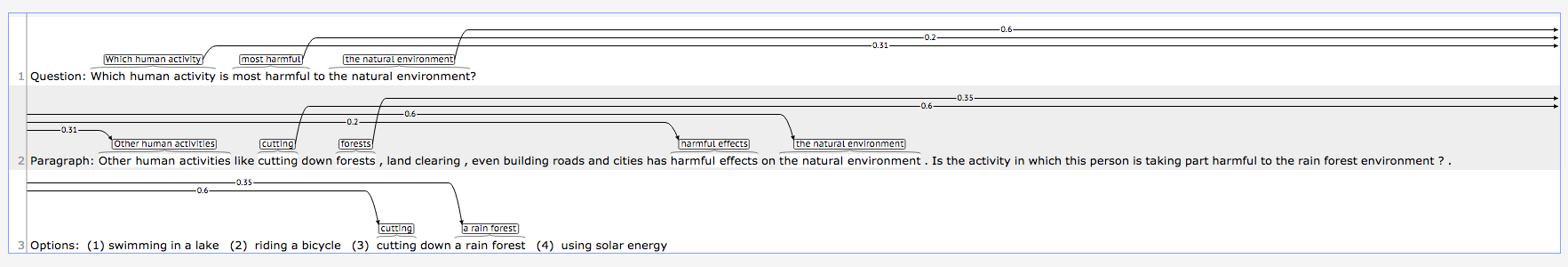}}
    \fbox{\includegraphics[trim=0.8cm 0.2cm 1cm 0.5cm, width=490pt,clip=true]{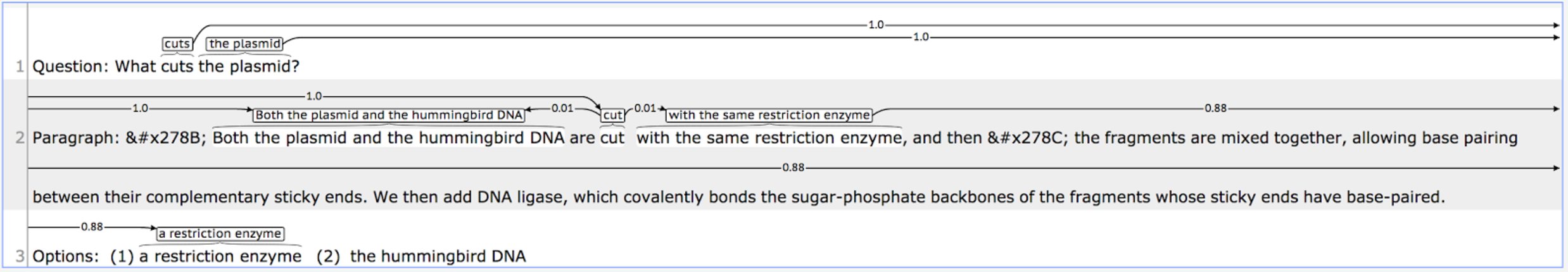}}
    \fbox{\includegraphics[trim=0.8cm 0.2cm 0.4cm 0.3cm, width=490pt,clip=true]{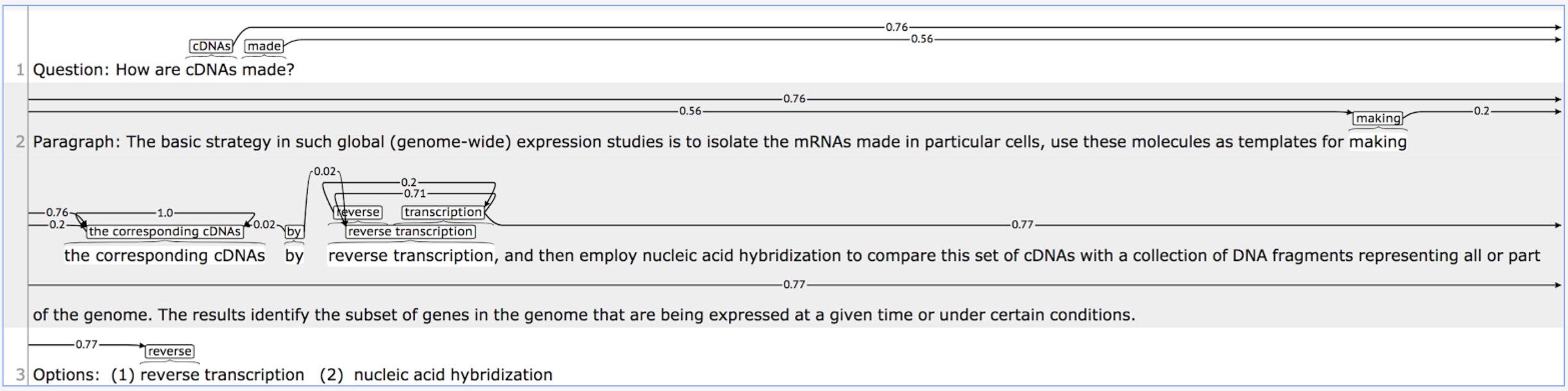}}
    \caption{Examples of system output for (1) top: Comb-1 (\simple\ alignment) (2) middle: Comb-2 (\verbSRL\ alignment) (3) bottom: Comb-5 (\verbSRL + \prepSRL\ alignment).  }
    \label{fig:output-simple-alignment}
\end{figure*}

This reasoning captures the least-constrained formulation for answering questions. We create alignments between \tokens\ view of the question, and spans of \shallowP\ view of the paragraph. Alignments are scored with the entailment module; the closer the surface strings are, the higher score their edge gets. 
There are variables to induce sparsity in the output; for example, penalty variables for using too many sentences in the paragraph, or using too many spans in the paragraph.  To Add more structure to this, we use \coref\ and \dep\ views of the paragraph; for example, alignments that are closer to each other according the dependency parse, get a higher score. The \coref\ links let the reasoning to jump in between sentences. Figure~\ref{fig:output-simple-alignment} shows an example output of this type of reasoning. As it can be seen, the alignment is using multiple terms in the questions and multiple spans in the paragraph, and the spans belong to one single sentence in the paragraph. 

\begin{figure*}[htb]
    \centering
    \resizebox{\linewidth}{!}{
    \fbox{
    \includegraphics[trim=1.0cm 0.7cm 0.8cm 0.4cm, clip=true,scale=0.315]{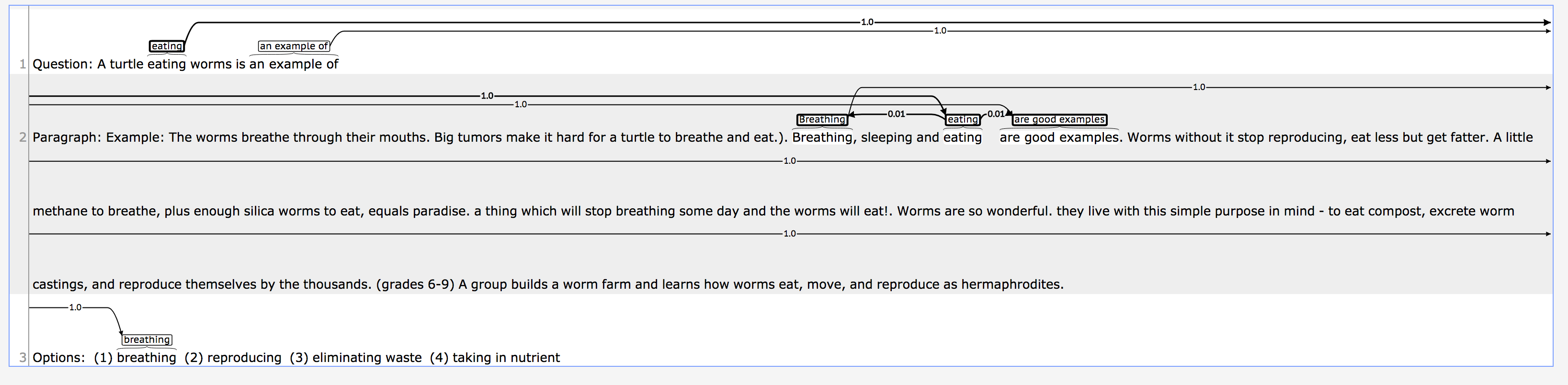}}
    }
    \caption{Example output of a question \textilp\ answered incorrectly due to a mistake in annotations. ``eating" in the paragraph is incorrectly recognized as a verb predicate, with ``breathing" as subject, resulting in this incorrect alignment. 
    }
    \label{fig:annotation-mistake}
\end{figure*}

\subsubsection{Comb-2 (\verbSRL\ Alignment)}
This reasoning is using \verbSRL\ views in both question and paragraph. The core of this alignment is creating connections between the predicate/arguments and the predicate/arguments of the paragraph, respectively. The edges are scored with our entailment system; hence the bigger the edge weight is, the higher chance of appearing in the output.  An example outputs are in Figure~\ref{fig:output-simple-alignment}. 

\subsubsection{Comb-5 (\verbSRL+\prepSRL\ Alignment)}
In this type of reasoning we observe the combination of \verbSRL\ and \prepSRL. This can be considered as an extension of Comb-2 (\verbSRL\ alignment), we the alignment is in between  \verbSRL\ in question, \verbSRL\ and \prepSRL\ in the paragraph. The arguments of the \verbSRL in the question are aligned to the arguments of either \verbSRL\ and \prepSRL\ in the paragraph. Predicates of the \verbSRL\ are aligned to the predicates \verbSRL. An example output prediction is shown in Figure~\ref{fig:output-simple-alignment}. 

Similar to the combinations explained, one can identify different combinations, such as Comb-3(\verbSRL+\coref\ Alignment), Comb-4(\verbSRL+\commaSRL\ Alignment) or \verbSRL+\nomSRL\ Alignment. 



\begin{table}[htb]
    \centering
    {\footnotesize 
    \begin{tabular}{L{7.5cm}}
        \hline 
            \ttitle{Variables: }  \\
            Active answer option variable \\ 
            Active answer option chunk variable \\ 
            \ttitle{Active variable constraints: }  \\
            If any edge connected to any active variable that belongs to answer option, the answer should be active \\
            If an answer is active, there must be at least one active edge connected to it. \\
            \ttitle{Consistency  constraints: }  \\
            Only one answer option should be active. \\
        \hline 
    \end{tabular}
    }
    \caption{Generic template used as part of each reasoning}
    \label{tab:general-body}
\end{table}

\begin{table}[htb]
    \centering
    {\footnotesize 
    \begin{tabular}{L{7.5cm}}
        \hline 
            \ttitle{Basic variables: }  \\
            Active question-terms \\ 
            Active paragraph-spans  \\ 
            Active paragraph sentence \\ 
            Question-term to paragraph-span alignment variable \\ 
            Paragraph-span alignment to answer-option term alignment variable \\
            \ttitle{Active variable constraints: }  \\
            Question-term should be active, if any edge connected to it is active.\\
            If a question-term is active, at least one edge connected to it should be active. \\
            Sentence should be active, if anything connected to it is active. \\
            If a sentence is active, at least one incoming edge to one of its terms/spans should be active. \\
            \ttitle{Question sparsity-inducing variables: }  \\
            More than $k$ active question-term penalty. (for $k = 1, 2, 3$) \\ 
            More than $k$ active alignments to each question-term penalty. (for $k = 1, 2, 3$) \\ 
            \ttitle{Paragraph sparsity-inducing variables: }  \\
            - Active sentence penalty variable. \\  
            \ttitle{Proximity-inducing variables: }  \\
            Active dependency-edge boost variable: if two variables are connected in dependency path, and are both active, this variable can be active. \\ 
            - Word-pair distance $\leq k$ words boost: variable between any two word-pair with distance less than $k$ and active if both ends are active. (for $k = 1, 2, 3$) \\ 
            \ttitle{Sparsity-inducing variables in answer options: }  \\
            - More than $k$ active chunks in the active answer-option. (for $k = 1, 2, 3$) \\ 
            - More than $k$ active edges connected to each chunk of the active answer option. (for $k = 1, 2, 3$) \\ 
        \hline 
    \end{tabular}
    }
    \caption{Comb-1 (\simple\ Alignment)}
    \label{table:basic-alignment}
\end{table}

\begin{table}[htb]
    \centering
    {\footnotesize 
    \begin{tabular}{L{7.5cm}}
        \hline 
            \ttitle{Variables: }  \\
            Active \verbSRL\ constituents in question (both predicates and arguments), $0.01$. \\  
            Active \verbSRL\ constituents in the paragraph (including predicates and arguments), with weight $0.01$.  \\ 
            Edge between question-\verbSRL-argument and paragraph-\verbSRL-argument (if the entailment score $ > 0.6$). \\ 
            Edge between question-\verbSRL-predicate and paragraph-\verbSRL-predicate (if the entailment score $ > 0.6$). \\ 
            Edge between paragraph-\verbSRL-argument and answer-options (if the entailment score $> 0.6$). \\ 
            Edges between predicate and its arguments argument, inside each \verbSRL\ frame in the paragraph, each edge with weight $0.01$. \\ 
            \\
            \ttitle{Consistency  constraints: }  \\
            Constraints to take care of active variables (e.g. edge variable is active iff its two nodes are active). \\
            At least $k = 2$ constituents in the question should be active. \\
            At least $k = 1$ predicates in the question should be active. \\ 
            At most $k = 1$ predicates in the paragraph should be active. \\ 
            For each \verbSRL\ frame in the paragraph, if the predicate is inactive, its arguments should be inactive as well.  \\
        \hline 
    \end{tabular}
    }
    \caption{Comb-2 (\verbSRL\ alignment)}
    \label{tab:verb-srl-reasoning}
\end{table}

\begin{table}[htb]
    \centering
    {\footnotesize 
    \begin{tabular}{L{7.5cm}}
        \hline 
            \ttitle{Variables: }  \\
            Active \verbSRL\ constituents in question (both predicates and arguments), with weight $0.001$. \\ 
            Active \verbSRL\ constituents in the paragraph (including predicates and arguments), with weight $0.001$. \\ 
            Active \prepSRL\ constituents in the paragraph (including predicates and arguments), with weight $0.001$. \\ 
            Edges between any pair of \prepSRL\ arguments in the paragraph and \verbSRL\ arguments, with weights extracted from \paragram (if these scores are  $> 0.7$). \\ 
            Edges between predicates and arguments of the \prepSRL\ frames, with weight $0.02$. \\ 
            Edges between predicates and arguments of the \verbSRL\ frames, with weight $0.01$. \\ 
            Edge between question-\verbSRL-argument and paragraph coref-constituents (if cell-cell entailment score $> 0.6$)  \\ 
            Edge between question-\verbSRL-predicate and paragraph \verbSRL-predicate (if phrase-sim score $> 0.5$)   \\ 
            Edge between paragraph \verbSRL-arguments and answer options (if cell-cell score is $> 0.7$) \\ 
            \\
             \ttitle{Consistency constraints:} \\ 
            Each \prepSRL\ argument has to have at least an incoming edge (in addition to the edge from its predicate) \\ 
            Not possible to have a \verbSRL\ argument (in the paragraph) \\  connected to two \prepSRL\ arguments of the same frame (no loop) \\ 
            Exactly one \prepSRL\ predicate in the paragraph \\ 
            At least one \verbSRL\ predicate in the paragraph \\ 
            At most one \verbSRL\ predicate  in the paragraph  \\ 
            Not more than one argument of a frame can be connected to the answer option \\ 
            Each \verbSRL\ argument in the paragraph should have at least two active edges connected to.  \\ 
        \hline 
    \end{tabular}
    }
    \caption{Comb-5 (\verbSRL+\prepSRL alignment)}
    \label{tab:verb-srl-prep-srl-reasoning}
\end{table}

\begin{table}[htb]
    \centering
    {\footnotesize 
    \begin{tabular}{L{7.5cm}}
        \hline 
            \ttitle{Variables: }  \\
            Active \verbSRL\ constituents in question (including predicates and arguments), with weight $0.001$. \\ 
            Active \verbSRL\ constituents in the paragraph (including predicates and arguments), with weight $0.001$. \\ 
            Active \coref\ constituents in the paragraph, with weight $0.001$. \\  
            Active \coref\ chains in the paragraph, with weight $-0.0001$. \\ 
            Edges between any pair of \coref-constituent in the paragraph that belong to the same \coref\ chain, with weight $0.02$. \\
            Edge between question-\verbSRL-argument and paragraph \coref-constituents (if entailment score $> 0.6$) \\ 
            Edge between question-\verbSRL-predicate and paragraph \verbSRL-predicate (if phrase-sim score $> 0.4$)   \\ 
            Edge between paragraph \verbSRL\ arguments and answer options (if symmetric entailment score is $> 0.65$) \\ 
            \\
            \ttitle{Consistency constraints:} \\ 
            Constraints to take care of active variables (e.g. edge variable is active iff its two nodes are active).  \\ 
            At most $k = 1$ \coref\ chain in the paragraph.  \\ 
            At least $k = 1$ constituents in the question should be active.  \\ 
            At most $k = 1$ \verbSRL\ predicates in the paragraph. (could relax this and have alignment between multiple SRL frames in the paragraph) \\ 
            The question constituents can be active, only if at least one of the edges connected to it is active.  \\ 
            Paragraph \verbSRL-predicate should have at least two active edges.\\ 
            If paragraph \verbSRL-predicate is inactive, the whole frame should be inactive. \\ 
        \hline 
    \end{tabular}
    }
    \caption{Comb-3 (\verbSRL+\coref\ alignment)}
    \label{tab:verb-srl-coref-reasoning}
\end{table}